%% file: paper_template.tex
\documentclass[letterpaper, 10 pt, conference]{ieeeconf}  
\IEEEoverridecommandlockouts                                  

\long\def\invis#1{}

\usepackage{multicol}
\usepackage{algorithm}
\usepackage{amsmath}
\usepackage{algpseudocode}
\usepackage{xcolor}
\usepackage{graphicx}
\usepackage{gensymb}
\usepackage{cite}

\newcommand{\NK}[1]{{\footnotesize\color{red}[{\bf NK:} \textsf{#1}]}}

\newcommand{\KJ}[1]{{\footnotesize\color{blue}[{\bf KJ:} \textsf{#1}]}}
\newcommand{\PRT}[1]{{\footnotesize\color{red}[{\bf PRT:} \textsf{#1}]}} 

\title{\LARGE \bf \textsc{UIVNav}: Underwater Information-driven Vision-based Navigation \\via Imitation Learning}

\author{Xiaomin Lin$^{1}$*, Nare Karapetyan$^{2}$*, Kaustubh Joshi$^{1}$, Tianchen Liu$^{1}$, Nikhil Chopra$^{1}$, \\Miao Yu$^{1}$, Pratap Tokekar$^{1}$, Yiannis Aloimonos$^{1}$
\thanks{This work was supported by USDA NIFA sustainable agriculture system program under award
number 20206801231805.}
\thanks{* Equal Contributors and Corresponding Authors.}
\thanks{$^{1}$Maryland Robotics Center (MRC), University of Maryland, College Park, MD 20742, USA. Emails: \texttt{\{xlin01, kjoshi, tianchen, nchopra, mmyu, tokekar, jyaloimo\}@umd.edu}.}
\thanks{$^{2}$Woods Hole Oceanographic Institution (WHOI), Woods Hole, MA 02543, USA. Emails: {\tt\small nare@whoi.edu}}
}

\begin{document}

\maketitle
\thispagestyle{empty}
\pagestyle{empty}


\input{sections/00_abstract}
\input{sections/01_intro}

\input{sections/02_related_work}

\input{sections/03_proposed_method}

\input{sections/04_experiments.tex}

\input{sections/05_conclusion.tex}

\bibliographystyle{IEEEtran}
\bibliography{references}

\end{document}

%% file: sections/00_abstract.tex
\begin{abstract}
Autonomous navigation in the underwater environment is challenging due to limited visibility, dynamic changes, and the lack of a cost-efficient, accurate localization system. We introduce \textsc{UIVNav}, a novel end-to-end underwater navigation solution designed to navigate robots over Objects of Interest (OOI) while avoiding obstacles, all without relying on localization. \textsc{UIVNav} utilizes imitation learning and draws inspiration from the navigation strategies employed by human divers, who do not rely on localization. \textsc{UIVNav} consists of the following phases: (1) generating an intermediate representation (IR) and (2) training the navigation policy based on human-labeled IR. By training the navigation policy on IR instead of raw data, the second phase is \textit{domain-invariant} --- the navigation policy does not need to be retrained if the domain or the OOI changes. We demonstrate this within simulation by deploying the same navigation policy to survey two distinct Objects of Interest (OOIs): oyster and rock reefs. We compared our method with complete coverage and random walk methods, showing that our approach is more efficient in gathering information for OOIs while avoiding obstacles. The results show that \textsc{UIVNav} chooses to visit the areas with larger area sizes of oysters or rocks with no prior information about the environment or localization. Moreover, a robot using \textsc{UIVNav} compared to complete coverage method surveys on average 36$\%$ more oysters when traveling the same distances. We also demonstrate the feasibility of real-time deployment of \textsc{UIVNav} in pool experiments with BlueROV underwater robot for surveying a bed of oyster shells.
\end{abstract}

%% file: sections/01_intro.tex
\section{Introduction}

Underwater navigation is a challenging task that has been the subject of numerous research studies~\cite{christensen2022recent}. In particular, interest in marine exploration and scientific sampling applications~\cite{hansen2018autonomous, karapetyan2021meander,manjanna2016efficient} has significantly increased the demand for efficient underwater navigation methods for exploration and data gathering. Many classical navigation methods developed for ground and aerial vehicles have been adapted to underwater environments~\cite{xanthidis2020navigation,jalal2021underwater}. Navigation in an underwater environment presents unique difficulties, such as limited visibility, dynamic environments, and a lack of cost-efficient, accurate localization systems. Underwater vehicles are commonly used for data collection to monitor environmental phenomena. To improve the efficiency of data collection, several complete coverage path planning methods have been developed~\cite{xu2014efficient,karapetyan2018multi} and adapted to underwater environments~\cite{vidal2019two, galceran2015coverage, xanthidis2021aquavis}. However, these approaches typically rely on partial or complete knowledge about the environment or assume the robot can localize itself. 

The absence of reliable localization~\cite{joshi2019experimental} requires new approaches for tackling underwater navigation or exploration problems. Towards this, vision-only based methods have been proposed for surveying coral reefs~\cite{manderson2018vision} and shipwrecks~\cite{karapetyan2021human}. Nevertheless, these approaches are domain-dependent as they have been trained over specific domains and objects and thus require retraining navigation policy each time a new Object of Interest (OOI) is being surveyed.

\begin{figure}[t]
    \centering
    {\includegraphics[width=\columnwidth]{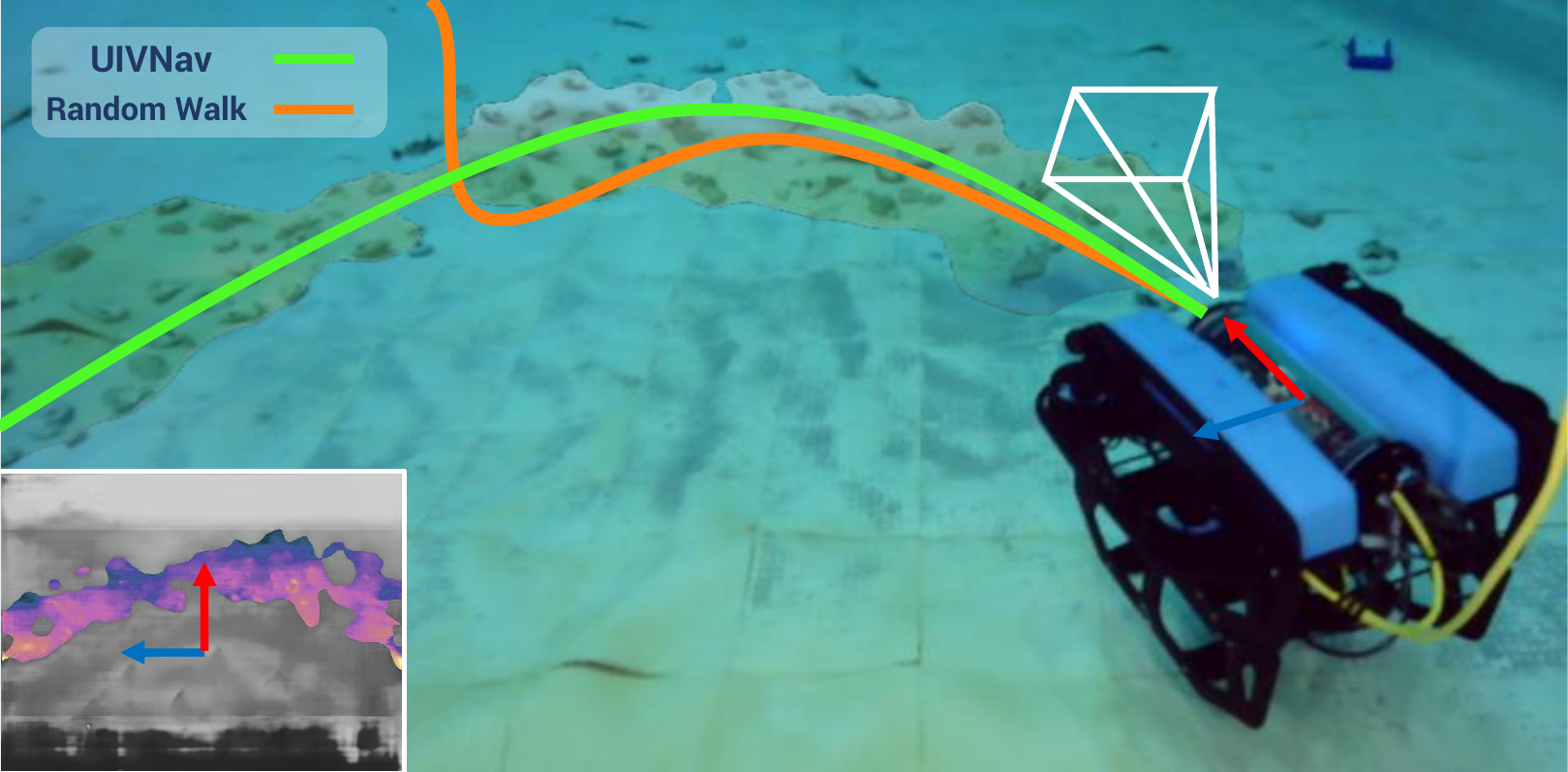}}
    \vspace{-3mm}
    \caption{Example of two trajectories of BlueROV surveying oyster-reef in a pool. On the bottom left is the current frame in the intermediate representation that the robot observes. \invis{\PRT{I would say add UINav and Radnom Walk along the green and red trajectories in the figure itself}}\invis{\textcolor{red}{Change the coloring - make red the Random Walk and green our method otherwise it's confusing. POOL PHOTO WOULD BE BETTER HERE}}}
    \vspace{-4mm}
    \label{fig:beauty_shot}
\end{figure}

We propose a new Underwater Information-driven Vision-based Navigation via Imitation Learning (\textsc{UIVNav}) method to address some of these challenges. Our method focuses on information gathering as the main objective --- we want to visit regions with OOI while avoiding obstacles. \textsc{UIVNav} is a type of imitation learning that uses only RGB camera information and generates an Intermediate Representation (IR) of each frame, which humans then use to label corresponding actions. IR allows us to train domain-invariant navigation policies to work across different OOI. Moreover, this method does not rely on localization but only on IR. We have compared our method with baseline navigation approaches -- random walk and complete coverage. The results indicate that \textsc{UIVNav} outperforms random walk and even surpasses localization-based complete coverage approach by surveying on average 36$\%$ more oysters when traveling the same distances. \invis{This demonstrates that our proposed method can also be combined with classical waypoint navigation methods and fill the gaps of data collection when the localization is inaccurate due to changes in environmental conditions.}

The main contributions of this work are: 
\begin{enumerate}
    \item The development of a domain-invariant navigation approach. This means that the navigation can gather information regardless of the type of domain being presented, reducing the need for retraining the navigation policy for new environments. We show this by deploying \textsc{UIVNav} on two different OOI oysters and large rock reefs, and in different domains the simulation and the pool. This provides a more generalized solution that can be applied in a wide range of underwater environments.
    \item A robot learns how to navigate successfully while covering (``seeing'') as much of the OOI as possible --- without relying on localization. This is an example of a new local navigation technique that would aid a class of global exploration and coverage algorithms. As navigation computational infrastructures (e.g., SLAM and related techniques) are reaching a high level of sophistication, the need arises to perform navigation and solve various tasks while doing so.
    \item Deployment of the proposed method in real-time in a large pool for surveying an oyster shell bed --- demonstrating the feasibility of real-world application of \textsc{UIVNav} (Fig.~\ref{fig:beauty_shot}).
\end{enumerate}

The rest of this paper is organized in the following manner. First, in Section~\ref{sec:related_work}, we discuss the recent advances in the field. The problem formulation and proposed approach are presented in Section~\ref{sec:approach}. Experimental validation in simulation and lab environment are discussed in Section~\ref{sec:experiments}. Finally, a conclusion with a discussion on possible future work is presented in Section~\ref{sec:conclusion}.

%% file: sections/02_related_work.tex
\section{Related Work}
\label{sec:related_work}

The problem of underwater navigation has been extensively studied in the field~\cite{christensen2022recent} and is a fundamental aspect of a variety of applications including underwater exploration, scientific sampling, or underwater photogrammetry. Navigation is typically defined as an optimization problem where the goal is to find the optimal path that minimizes some cost function --- obstacle avoidance~\cite{braginsky2016obstacle}, energy cost~\cite{yang2021energy}, etc. Some classical planning methods have been adapted to underwater domains as well. ~\cite{xanthidis2020navigation} presents an augmentation of a classical optimization package for manipulators, to perform 3D trajectory planning of an underwater vehicle that uses either a known map or an online constructed local map. AquaVis~\cite{xanthidis2021aquavis}, an active perception system, was proposed to enhance this work by incorporating visual inspection. It identifies visual goals in real-time, moves towards the target location, and enhances the view of nearby visual goals from the preferred distance. These methods have been shown to have promising results, but they assume the existence of precise localization.

P{\'e}rez-Alcocer et al.~\cite{perez2016vision} developed a robust vision-based navigation system in an underwater environment with limited visibility. Manzanilla et al.~\cite{manzanilla2019autonomous} employed a vision-based localization approach using an Extended Kalman Filter (EKF), and designed a proportional integral derivative (PID) controller for the autonomous navigation of an underwater vehicle. Bobkov et al.~\cite{bobkov2016adaptive} proposed an adaptive method using stereo camera data to improve navigation performance. These methods attempted to incorporate planning with state estimation but they do not consider efficient data collection as an objective.

Very few attempts have been made in the literature to address underwater coverage and data acquisition problems when operating in 3D space. Extending on classical coverage methods,~\cite{galceran2012efficient} suggests segmenting the environment based on similar depths and in later work~\cite{galceran2015coverage} takes into account state estimation uncertainties. ~\cite{vidal2017online} proposes a next-best view approach but it is significantly constrained by the certainty of the state estimation. To overcome the complexity of 3D exploration, a simplified formulation of the problem is considered, such as 2D mapping of underwater structure \cite{vidal2019two}. These methods are attempts towards 3D exploration but they are also limited by the prior information about the environments given in a map form or by the state estimation accuracy.

To overcome limitations of reliable state estimation while collecting underwater data~\cite{manderson2018vision} and ~\cite{karapetyan2021human} have proposed an end-to-end vision-based approach that surveys highly unstructured environments such as coral reefs and shipwrecks. In these works, the navigation policy is learned through supervised imitation learning. The human diver is asked to label image sequences with yaw and pitch changes such that the resulting action will drive the robot towards coral/shipwreck and simultaneously avoid obstacles. \cite{Manderson2020rss} extended these works to incorporate user-defined goal conditions, while \cite{liu2022learning} adapted it to navigation in confined areas. Each of these methods is specifically trained for exploring a particular ROI - coral reef or shipwreck and will require new labeling and retraining for navigation over new ROI.

In summary, various methods have been proposed for path planning in underwater environments, with varying degrees of dependence on prior information and the complexity of the environment. The challenge remains to develop a general solution that can navigate effectively in unknown environments while avoiding obstacles and performing efficient data collection.\invis{ maximizing data collection with minimal retraining requirements.} Our work is motivated by human diver-inspired coverage and navigation methods~\cite{karapetyan2021human, manderson2020vision, smolyanskiy2017toward}. We introduce a novel domain-invariant navigation policy strategy that using IR reduces policy retraining and labeling overhead. Our approach provides a more generalized solution that is less sensitive to changes in the environment and can be used when the robot is unable to localize, whether in combination with global planning methods or standalone.

%% file: sections/03_proposed_method.tex
\section{Approach}
\label{sec:approach}

\begin{figure*}[ht!]
\vspace{3mm}
\includegraphics[width=0.98\textwidth]{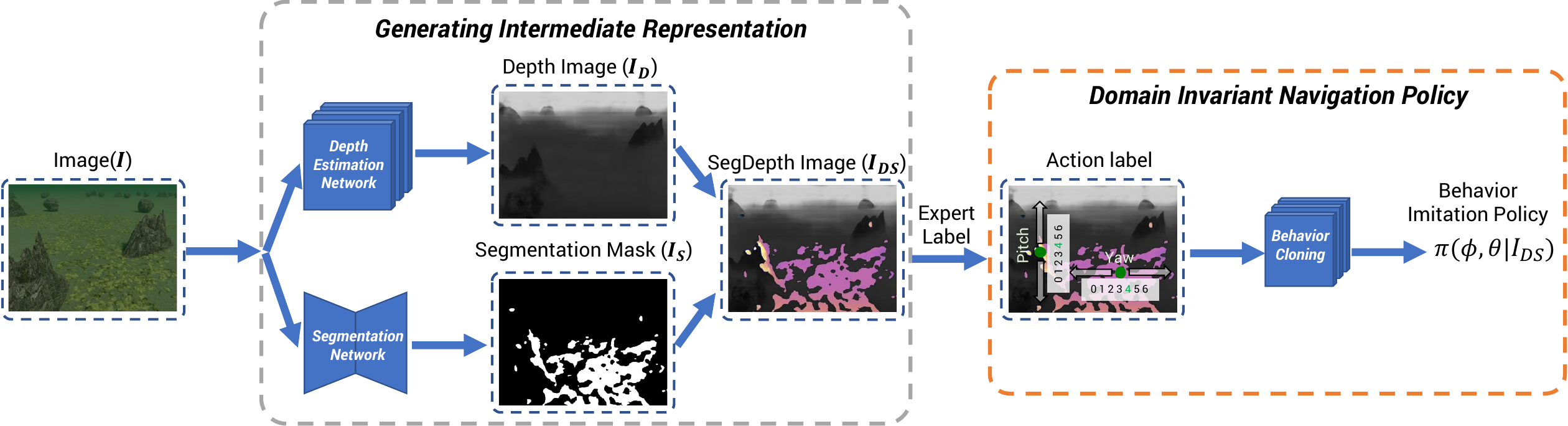}
\centering
\vspace{-3mm}
\caption{An overview of our approach. \textsc{UIVNav} consists of two phases: Generating Intermediate Representation (gray box) and Data Invariant Navigation Policy (orange box). Image ($I_D$) is fed into the depth estimation network and the segmentation network. The output from the depth estimation network ($I_D$) and the output from the segmentation network ($I_S$) are used to generate the intermediate representation ($I_{DS}$). This is fed into an imitation learning network to train a navigation policy that predicts yaw and pitch changes. Due to training on ($I_{DS}$) the navigation policy is domain invariant and will not require restraining nor labeling with new OOI.}
\vspace{-5mm}
\label{fig:overview}
\end{figure*}

Our method, \textsc{UIVNav}, is designed to maximize data collection for OOI and avoid obstacles without relying on localization.
Our proposed method is based on imitation learning and is influenced by intuition used by human divers who do not rely on localization when collecting data. The backbone of the \textsc{UIVNav} is the intuition that when using only vision-based navigation to collect more data and avoid obstacles, two important cues are intrinsically used by humans: (1) recognition of the OOI --- segmentation of the OOI, and (2) spatial understanding of proximity of the obstacles or OOI --- depth estimation. We use this intuition to design a framework that achieves this objective. \invis{A main advantage of our method is the use of IR for training navigation policy which makes it a domain invariant policy, avoiding the need for re-labeling and re-training for different OOI.}

An overview of the proposed approach is illustrated in Fig.~\ref{fig:overview}. Our two-stage system consists of \textit{Generating Intermediate Representation} and \textit{Domain Invariant Navigation Policy}. In the first stage, the robot's images ($I$) are fed into the depth estimation network and the segmentation network simultaneously. The outputs from the depth estimation network ($I_D$) and the outputs from the segmentation network ($I_S$) are used to generate the intermediate representation ($I_{DS}$). Creating the intermediate presentation ($I_{DS}$) aims to eliminate domain-specific information and allow the system to gather information for different OOI while avoiding obstacles. We assume that $I_S$ and $I_D$ are available or can be generated using existing methods~\cite{kirillov2023segment,lin2023oysternet,ronneberger2015u,yu2023udepth}.

In the second stage, the Domain Invariant Control Policy phase, we label the intermediate presentation frames ($I_{DS}$ from stage one) with actions (pitch and yaw changes) for obstacle avoidance and information gathering.
With the labeled $I_{DS}$, we train a behavior cloning network that controls the robot to maximize the information gain for the OOI while avoiding obstacles. We present each stage in the following subsections.

\subsection{Intermediate Representation}
The networks that generate the intermediate representation must be altered with other state-of-the-art networks for various applications. For different domain work (air, ground and oceanic), and tasks (coral monitoring, aerial forester monitoring), the dataset used to train those networks shall also be different. We will briefly go through the networks and datasets we used and then define our intermediate representation in this subsection.
\subsubsection{Depth Estimation Network}
We need to acquire a corresponding dataset to train a depth estimation network in the underwater oyster reef environments. We utilized a convenient oyster simulation environment, OysterSim~\cite{lin2022oystersim}, to create the dataset consisting of 2000 image pairs of RGB images and the corresponding depth images(ground truth from the simulation). 
We adopted the monocular depth estimation network from Keras~\cite{chollet2015keras}, with Unet~\cite{ronneberger2015u} backbone. We used Inverse Huber Loss~\cite{laina2016deeper} for the depth estimation network. Sample output from the depth estimation network ($I_D$) can be seen in Fig.~\ref{fig:overview}.
\subsubsection{Segmentation Network}
Similarly, we utilized OysterSim~\cite{lin2022oystersim} to create the dataset of 2000 image pairs of RGB images of oyster reefs and the corresponding segmentation masks(ground truth from the simulation).  
We used the state-of-the-art oyster detection network, Oysternet~\cite{lin2023oysternet}, for segmenting the OOI. Sample output from the segmentation network ($I_D$) can also be seen in Fig.~\ref{fig:overview}.

We want to generate SegDepth Image ($I_{DS}$) for all images ($I$) in the Dataset ($\mathcal{D}$). Depth image ($I_D$) is a grayscale image with only 1 channel. We first want to save only the depth information for the OOI ($I_{D\_OOI}$) and only the depth information for non-OOI ($I_{D\_OOI}^\prime$). We then used the color mapping function ($M$) from \texttt{matplotlib}~\cite{Hunter:2007} that maps the 1-channel $I_{D_{OOI}}$ to 3-channel (RGB) $I_{DS_{OOI}}$. The depth information is still preserved in the 3-channel image and its color (3-channel information) represents the region of interest. We stack (repeat) the $I_{D_{OOI}}^{\prime}$ three times to get the 3-channel depth information ($I_{DS_{OOI}}^\prime$) for non-OOI. Finally, we combine these two 3-channel information to get our SegDepth Image ($I_{DS}$). 
\begin{algorithm}
\caption{Generating SegDepth Image}\label{alg:cap}
\begin{algorithmic}[1]
\For{ $I \in \mathcal{D}$}
    \State \texttt{$I_{D_{OOI}}=I_D \ \& \ I_S $}
    \State \texttt{$I_{D_{OOI}}^\prime=I_D\ \& \ \neg I_S$}
    \State \texttt{$I_{DS_{OOI}}=M(I_{D_{OOI}}) $}  \Comment{Mapping function}
    \State \texttt{$I_{DS_{OOI}}^\prime = stack(I_{D_{OOI}}) $} \Comment{stack 1-channel 3 times}
    \State \texttt{$I_{DS}= I_{DS_{OOI}}\ + \ (I_{DS_{OOI}}^\prime) $}
\EndFor
\State \textbf{return} $I_{DS}$
\end{algorithmic}
\end{algorithm}
\subsection{Domain Invariant Navigation Policy}
To learn our Domain invariant navigation policy, \textsc{the UIVNav} approach follows a similar pipeline as the behavioral cloning method presented in the literature~\cite{karapetyan2021human,liu2022learning,manderson2018vision,smolyanskiy2017toward}. The major distinctions are the navigation policy input of the network and the labeling objectives. We use the Intermediate Representation ($I_{DS}$) instead of the RGB image as our input data for imitation learning. 
We assume the robot is positioned at least 5m above the OOI, has a front-facing camera and the environment contains obstacles that can exceed 5m height. This is a typical case for top-view data collection common in oyster, coral, or forestry monitoring applications. We generate ($I_{DS}$) on the raw images collected from such an environment to perform the navigation policy labeling. A human diver performs the labeling through a Graphical User Interface (GUI) that shows the current frame and possible classes for yaw and pitch changes (see Action Label in Fig.~\ref{fig:overview}). 

The labeling objectives of the human diver that collects data over OOI are as follows: (1) prefer to move towards OOI; (2) move towards larger areas of the OOI that are further away; (3) avoid obstacles. Yaw and pitch values are each encoded as a discrete set of seven class labels similar to the related works suggested in the literature~\cite{manderson2018vision,karapetyan2021human}. A Convolutional Neural Network (CNN) with a ResNet-50 backbone is trained on 15000 labeled input images ($I_{DS}$) to predict these relative heading changes in yaw and pitch. 

We denote the $f_{\theta}(I_{DS})$ and $f_{\phi}(I_{DS})$ for the categorical distribution prediction (vector of size 1 by 7 for the probability of each class) for yaw and pitch respectively with the input of $I_{DS}$ from the behavior imitation policy. There are seven class prediction labels with a mean of 3 for the yaw and pitch which are described as $ C_{\theta}$ and $C_{\phi}  in \{0,1,2,3,4,5,6\}$. $C_{\theta}$ corresponds to desired yaw changes of $\{3\delta_{\theta}, 2\delta_{\theta}, \delta_{\theta}, 0, -\delta_{\theta}, -2\delta_{\theta}, -3\delta_{\theta}\}$ with respect to the current image frame. Here the positive change represents a clockwise heading change and the negative change represents a counterclockwise heading change with the amount of $\{\delta_{\theta}, 2\delta_{\theta}, 3\delta_{\theta}\}$. Similarly $C_{\phi}$ to the desired pitch changes with respect to the current image frame. Here the positive change represents an upward-heading change.
\invis{From the navigation in the forester trails and the dense coral, Seven classes were selected because they were proven to provide adequate fidelity in control actions while still being a manageable set for machine learning.} 

When applying the behavior imitation policy, we choose $\delta_{\theta}=5^{\circ},\delta_{\phi}=5^{\circ}$ as sharp changes are undesirable. As suggested by \cite{smolyanskiy2017toward}, we also performed label smoothing on the ground truth label. We treat  $f_{\theta}(I_{DS})$ and $f_{\phi}(I_{DS})$ separately as we are training the policy while using the same loss function below. As we believe the wrong prediction for one axis shall not penalize the right prediction for the other axis.  The following loss is used to train the policy $\pi(\phi,\theta|I_{DS})$: 
\begin{equation}
\mathcal{L}_{loss}(\mathcal{D},w)=\lambda*\mathcal{L}_{CCE}(\mathcal{D},w)+(1-\lambda)*\mathcal{L}_{KL}(\mathcal{D},w)
\end{equation}
The $\lambda$ here is a weight factor for the loss function. We choose $\lambda=0.1$. $\mathcal{L}_{CCE}$ is categorical cross-entropy loss and $\mathcal{L}_{KL}(\mathcal{D},w)$ is KL divergence from \cite{joyce2011kullback}.

The resulting predicted 7-class discrete network values for yaw and pitch are transformed into low-level actuator commands that the robot's motion controllers carry out during inference.

%% file: sections/04_experiments.tex
\section{Experimental Evaluation}
\label{sec:experiments}
\begin{figure}[t!]
\centering
\vspace{2mm}
\includegraphics[width=0.45\textwidth]{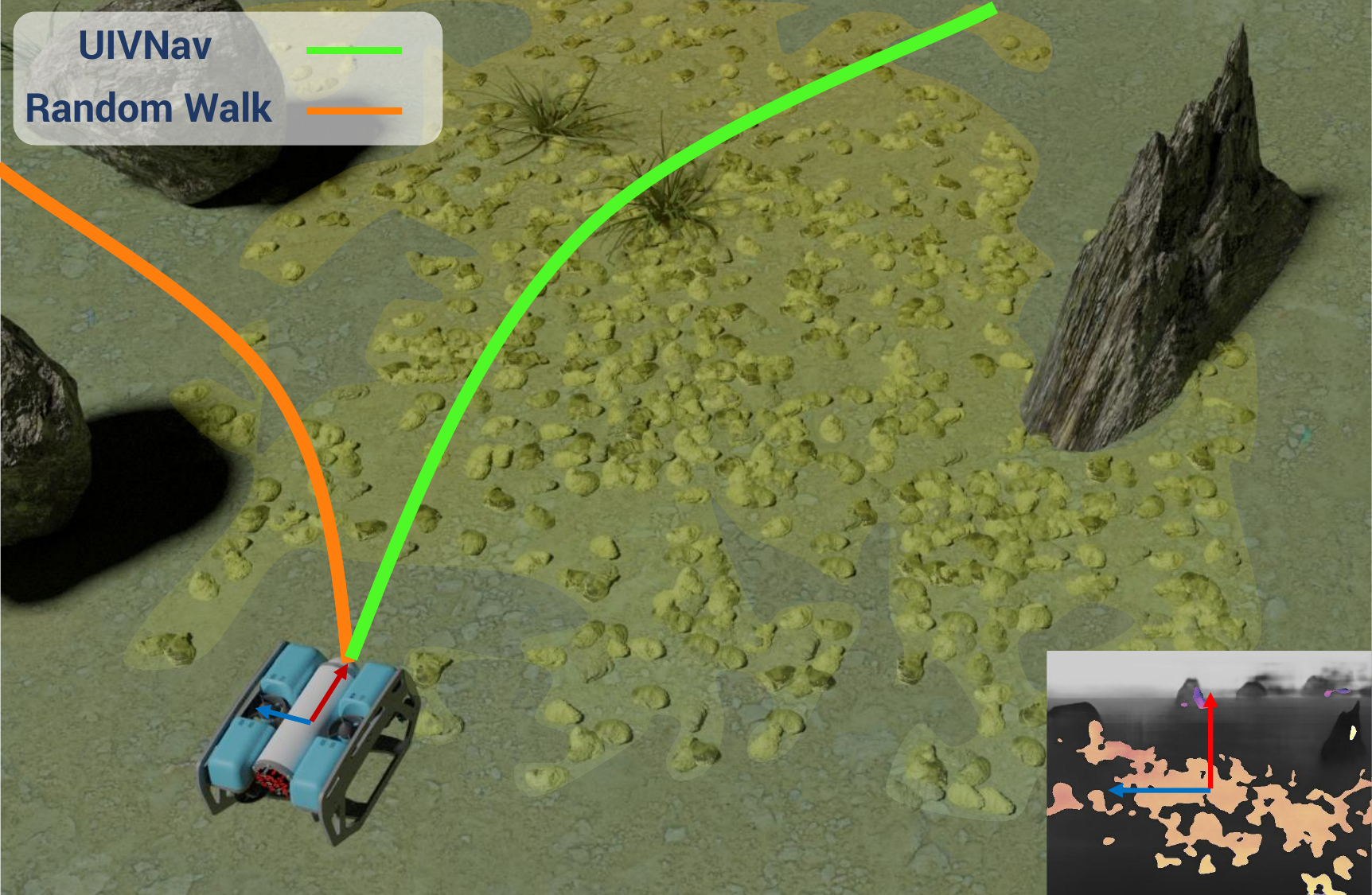}
\vspace{-2mm}
\caption{Example of BlueROV surveying reef of oysters in simulation (translucent mask is applied on the oyster reef for better visualization.)  } 
\vspace{-5mm}
\label{fig:bluerov_in_sim}
\end{figure}
We use the oyster reef monitoring application as a case study for demonstrating the capabilities of~\textsc{UIVNav}. We performed extensive quantitative and qualitative comparisons of our method with two baseline methods --- (1) Brownian Bridge (BB)~\cite{chow2009brownian} random walk and (2) an offline complete coverage method based on Boustrophedon Cell Decomposition (BCD)~\cite{xu2014efficient}. Brownian Bridge-based motions have been used to study animal trails~\cite{horne2007analyzing} and can have the potential to replicate the distribution of natural phenomena such as oyster or coral reef growth patterns. It has also been widely used in exploration tasks~\cite{pang2019swarm, wagner1998robotic}. As our goal is to maximize the collected data of OOI, we compare our method to the complete coverage path planning method, which is the state-of-the-art approach for completely covering the area when the prior map is given. In practice, farmers or scientists will perform a lawnmower pattern --- naive version of BCD pattern to navigate their vehicles and collect data or dredge oysters. For the BCD algorithm, we only assume the boundaries of the region and obstacle locations are given but not the distribution of the OOI. Note that we do not use any prior information about the environment when running \textsc{UIVNav}.  

In this section, we will first conduct an evaluation in a simulated environment with oysters. Additionally, to demonstrate the generalizability of our method to navigate over different OOI, we test in a rock reef environment. Finally, we show results from a deployment of \textsc{UIVNav} on a BueROV underwater robot in a \invis{Neutral Buoyancy} pool to assess the real-world viability of our approach in real-time.

\subsection{Simulation}
\invis{\textcolor{teal}{\NK{NOT SURE IF THIS paragraph IS NESSESARY - need to discuss this}We conducted extensive testing and validation in simulated environments. 15,000 training samples and 5,000 validation samples composed of images, depth ground truth, and oyster segmentation ground truth are rendered in the simulation. \textcolor{red}{HAND LABELED INTERMEDIATE REPRESENTATION? Based on the intermediate representation we computed, we manually labeled this dataset of pitch and yaw labels for training as discussed in section~\ref{sec:approach}.} We trained Resent-50 as our backbone with a learning rate of 0.001 coupled with Adam Optimizer. We used a weighted loss which is explained in Section~\ref{sec:approach}. For detecting the OOI, we trained OysterNet\cite{lin2022oysternet} with generated images and oyster segmentation ground using a learning rate of 0.001 with decay. We use the Adam optimizer coupled with the Jaccard loss function. For the depth estimation network, we trained a Unet\cite{ronneberger2015u} with generated images and depth ground using a learning rate of 0.001. We used a batch size of 32 and trained the network for 100 epochs.}
}
\begin{figure*}[t!]
\vspace{5mm}
\begin{tabular}{c}
\includegraphics[width=0.97\textwidth]{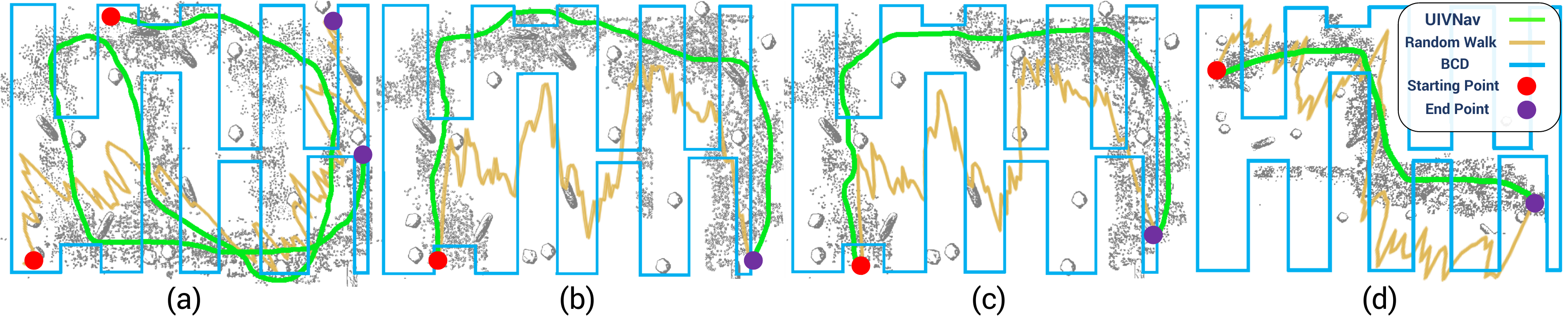}
\label{fig:S4_bm}
\end{tabular}
\vspace{-3mm}
\caption{Comparing various exploration strategies across four oyster reef patterns-- (a) Grid world, (b) E-shape, (c) Disconnected Paths, (d) Branching Corridor.} 
\vspace{-5mm}
\label{fig:all}
\end{figure*}
\begin{figure}[t!]
\centering
\includegraphics[width=0.46\textwidth]{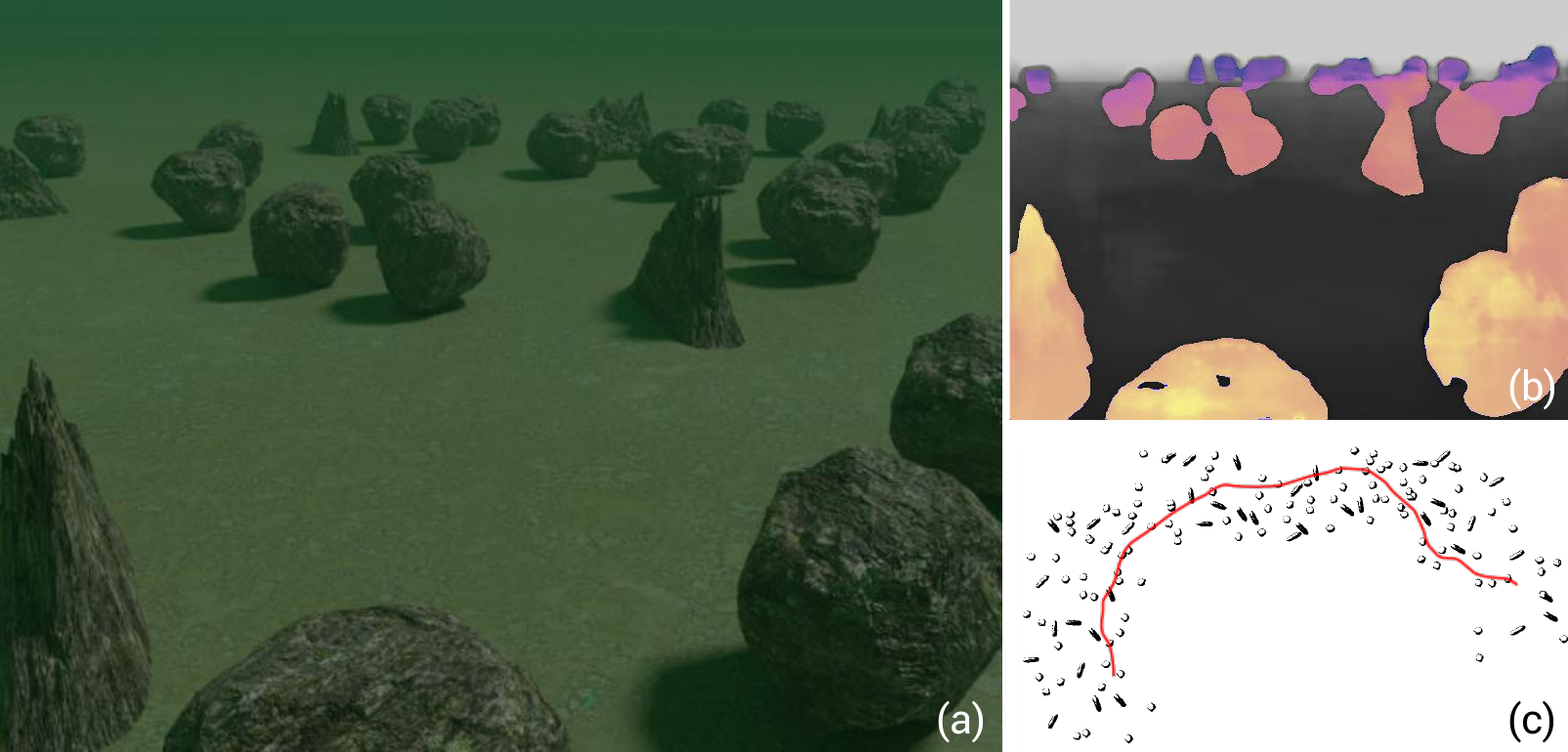}
\vspace{-3mm}
\caption{\textsc{UIVNav} surveying reef of rocks OOI with no additional human-expert labeling or training. (a) The current frame of the robot field of view in OysterSim simulation. (b) Intermediate representation of the current frame that the robot uses to make navigation decisions. (c) The complete trajectory over a rock reef.} 
\vspace{-5mm}
\label{fig:rocks}
\end{figure}
To generate underwater oyster and rock reef environments we used an open-source oyster simulation framework based on Blender~\cite{blender} game engine, OysterSim~\cite{lin2022oystersim}. We use a model of the BlueROV robot in the simulation as shown in Fig~\ref{fig:bluerov_in_sim}. We have generated different possible distributions of oyster reefs and one sample of rock reef. When designing the scenarios we took into account that underwater natural reefs -- oyster or coral --- usually do not grow separately but they tend to be in large groups. Nevertheless, they can have different discontinuous patches. As such, our test scenarios consist of the following instances:
\begin{enumerate}
    \item Grid World - a reef with connected continuous patches of oysters with multiple branches (Fig.~\ref{fig:all}a).
    \item E-shape - C-shaped wide oyster reef patch with a narrow oyster patch in the middle (Fig.~\ref{fig:all}b).
    \item Disconnected Paths -  a reef consisting of two large oyster groups that are separated by a large sandy area~(Fig.~\ref{fig:all}c).
    \item Branching Corridor - a reef with wide and narrow oyster patches branching from the opposite sides of the reef (Fig.~\ref{fig:all}d).
    \item Rock Reef - a reef of large rocks with continuous C-shape (Fig.~\ref{fig:rocks}c).
\end{enumerate}
\vspace{-2mm}
\subsection{Results}
\begin{figure*}[t!]
\begin{tabular}{ccc}
\includegraphics[width=0.34\textwidth]{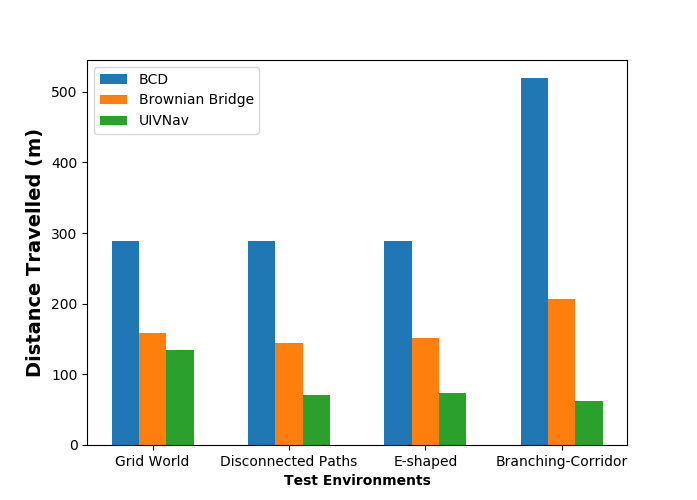}
\label{fig:S4_uvnav}& \hspace{-1cm}
\includegraphics[width=0.34\textwidth]{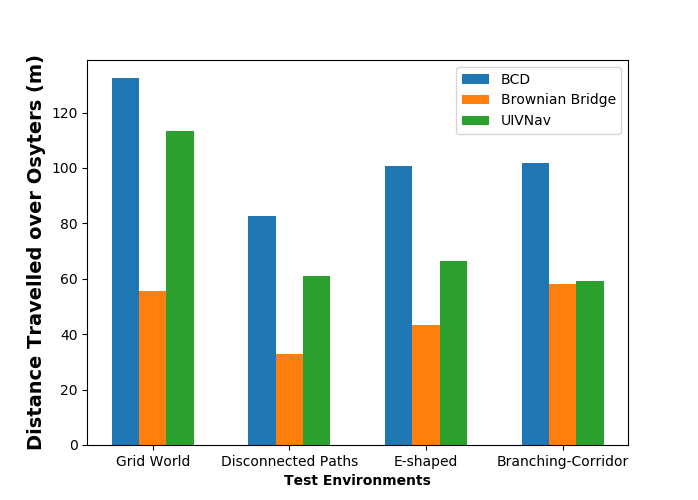}
\label{fig:S4_bc}&\hspace{-1cm}
\includegraphics[width=0.34\textwidth]{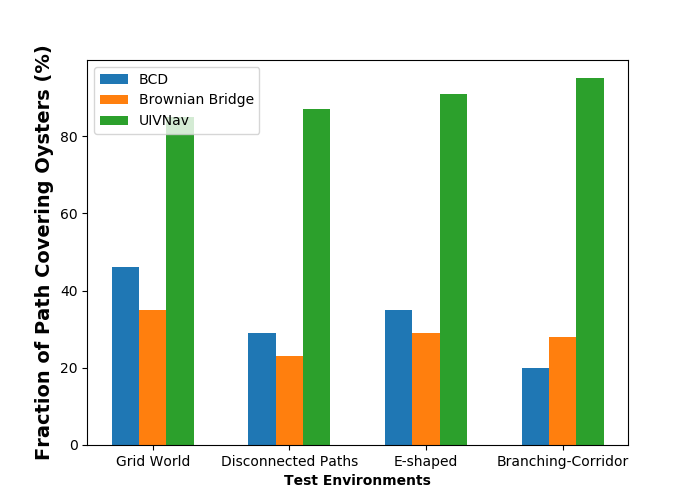} 
\label{fig:S4_bm}
\end{tabular}
\caption{Quantitative Evaluation of Different Strategies on Different Environments. BB method travels longer distances moving on fewer areas with oysters compared to~\textsc{UIVNav}. Even though compared to BCD, the~\textsc{UIVNav} method collects on average 29$\%$ less oyster data it travels 70$\%$ less distance. This shows that~\textsc{UIVNav} trajectories survey on average 36$\%$ more oysters than the BCD method when traveling the same distances.} 
\label{fig:results}
\end{figure*}

Since our objective is to maximize the data collected we have measured \textit{Distance Traveled} with \textit{Distance Traveled over Oysters} in meters and calculated the \textit{Percentage of Oysters} contained within each generated path assuming the robot travels at a consistent speed. As depicted in  Fig.~\ref{fig:results}, our results prove the hypothesis that our method~\textsc{UIVNav} is better than the random walk. Compared to the complete coverage method, the results indicate that a robot using~\textsc{UIVNav} collects on average 29$\%$ less oyster data while traveling on average 70$\%$ less distance. This shows that ~\textsc{UIVNav} trajectories survey on average 36$\%$ more oysters than the complete coverage method when traveling the same distance.

In each of the four test environments, we show that~\textsc{UIVNav}  makes qualitatively more efficient decisions (Fig.~\ref{fig:all}). In \textit{Grid World}, \textsc{UIVNav}  almost completely covers the oyster reef without navigating over sand patches and avoids all of the obstacles (Fig.~\ref{fig:all}{a}). In \textit{E-shape}, we can observe that \textsc{UIVNav} chooses to turn on a wider region towards the endpoint rather than traveling through the middle patch --- demonstrating the ability to infer "future reward (Fig.~\ref{fig:all}{b}). \textit{Disconnected Paths} scenario, \textsc{UIVNav} is able to navigate through the gap and locate a new OOI in sight, enabling it to effectively survey over the remaining portion of the oyster reef (Fig.~\ref{fig:all}{c}). The last scenario with oyster reef --- \textit{Branching Corridor}, demonstrates \textsc{UIVNav}'s ability to select wider regions with oyster surface to continue exploration (Fig.~\ref{fig:all}{d}). 

We also qualitatively evaluate\textsc{UIVNav} on the \textit{Rock Reef} scenario to show that the navigation policy is domain invariant. To obtain the intermediate representation for different OOI, we re-train the segmentation network for detecting the rocks. On a newly generated IR, we directly apply a trained navigation policy. Robot successfully explores new OOI without the need for additional human-expert labeling or re-training of the policy (Fig.~\ref{fig:rocks}).
\invis{
\begin{figure}[t!]
    \centering
    {\includegraphics[width=0.48\textwidth]{./figures/top_view.png}}
 \vspace{-5mm}
	\caption{BlueROV2 Deployed in a Neutral Buoyancy Pool with Yaw and Pitch control with E-shape oyster shells distributed at the bottom. }
	\label{fig:top_view}
\vspace{-5mm}
\end{figure}
}
\begin{figure*}
\begin{tabular}{c}
\includegraphics[width=0.97\textwidth]{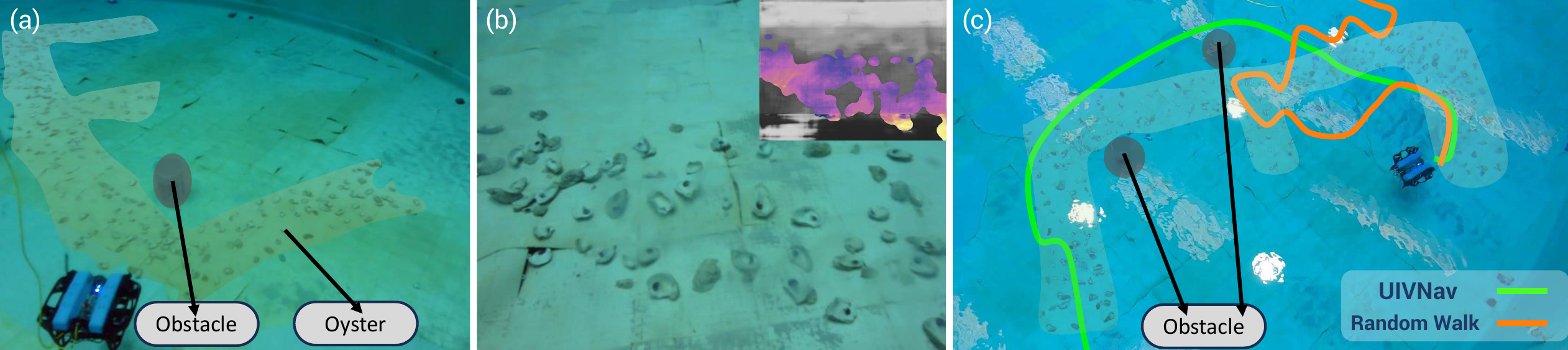}
\label{fig:S4_bm}
\end{tabular}
\caption{(a) BlueROV2 Deployed in a Neutral Buoyancy Pool with Yaw and Pitch control with E-shape oyster shells distributed at the bottom. (b) BlueROV's field of view.\invis{\NK{the robot's view can go here}} (c) Experimental result of two trajectories of BlueROV surveying oyster shells in a pool. \invis{\KJ{Should we add an image with the intermediate representation as well here?}\NK{I would add on (c) on the right bottom corner the same way the label for the trajectories as it is right now in Fig 1}}} 
\vspace{-5mm}
\label{fig:pool_experiment}
\end{figure*}
\subsection{Lab Deployment}
\subsubsection{Hardware Setup}We conducted pool experiments with the BlueROV2 robot in Heavy Configuration \cite{robotics2016bluerov2} with 8 thrusters for motion in all 6 degrees of freedom. The BlueROV2 is equipped with the following onboard sensors: TDK Invensense ICM-20602 Inertial Measurement Unit (IMU), Waterlinked DVL-A50 Doppler Velocity Logger (DVL), and a Bosch BMP280 pressure sensor, an onboard Low Light HD USB Camera with 1080p resolution. \invis{horizontal field of view 80\degree and vertical 64\degree. Light sensitivity 0.01 lux. }The robot works on the \textit{Navigator} flight controller with ArduSub as the software suite. \invis{MAVLink protocols are used for interfacing the BlueROV2, using the~\texttt{pymavlink} library. }The BlueROV2 is controlled by giving Pulse-Width Modulation (PWM) to the thrusters through I2C interface PCA9685 --- a 16-Channel PWM Module. 

The BlueROV2 is connected through the tether to the Ground Central Station (GCS) - Dell Precision 7560. The GCS is equipped with the NVIDIA RTX A2000 GPU.

\subsubsection{Software Setup} 
All drivers for processing the data and autonomous control of the robot are ROS-based software written in Python. The \textsc{UIVNav} was deployed on GCS, executing navigation commands in real-time at a frequency of 0.5Hz.
The commands generated from the \textsc{UIVNav} network are mapped to the PWM inputs for thrusters to change the BlueROV2's pitch, yaw, surge, and heave. The network gives us a set of commands ($c$) as single digit numbers, which determine the desired velocity ($\nu$) --- ($\nu = c * \alpha$),
where $\alpha$ is a constant determining factor by which the magnitude of speed changes. The input value for the PWM input is calculated as follows:
\begin{equation}
\vspace{-2mm}
\text{PWM}_\eta = 1500\mu\text{s} + k_\eta \nu
\vspace{-0.5mm}
\end{equation}
where 1500$\mu\text{s}$ keeps the thrusters at neutral position, $\eta$ denotes the concerned degree of freedom and $k$ denotes the gain for each direction which is calculated experimentally.

\invis{\subsubsection{Intermediate Representation}The main assumption in \textsc{UIVNav} is the availability of segmentation and depth masks for generating the intermediate representation. Since the lab setup (Fig.~\ref{fig:pool_experiment}) is very different from the simulated oyster reefs, we retrained the detection network for OOI. We first manually collect 300 images of the oysters from the bottom of the pool and label their segmentation masks. These images were trained with the OysterNet\cite{lin2023oysternet} model for detecting OOI with a learning rate of 0.001 with decay. On the resulting intermediate representation without any additional training the \textsc{UIVNav} has been executed to produce navigation policy.\invis{We also use the Adam optimizer along with the Jaccard loss function. The actions involve a set of pitch, yaw, and surge commands. Additionally, the BlueROV2 was programmed to hold its depth at 1 ft above the pool bottom using the feedback of its pressure sensors. Notice that, we did not retrain our navigation policy and depth estimation network. We only re-train the detection network for OOI here. }}

\subsubsection{Experiment setup} We distributed up to 300 oyster shells at the bottom of the neutral buoyancy tank. The neutral buoyancy tank is 50 feet across and 25 feet deep. The E-shape pattern of the oyster shell patch is bounded by (Fig.~\ref{fig:pool_experiment}{a}) a 40x20 feet area --- which occupies half of the tank. The breadth of the oyster patch averages 1 to 2 feet, which fits nearly 3-5 shells.

The real-world deployment of the robot posed several challenges that significantly increased the task's difficulty. Namely, the robot's motion was influenced by the altitude at which front camera images were captured, drift from turning and surging, and external forces that affected the robot's turning radius and positioning. Nevertheless, the robot successfully surveyed the oyster shells using the navigation policy derived from our method (see Fig.~\ref{fig:pool_experiment}{c}). \invis{In comparison as a baseline, we showed a random walk in Fig.~\ref{fig:pool_experiment}{c} in yellow.}

In this experiment, without prior knowledge of the environment and localization, the \textsc{UIVNav} moves the robot towards oyster reefs (Fig.~\ref{fig:pool_experiment}{b)}--- maximizing the information gain over OOI while avoiding two obstacles (two water buckets). By using random walk, the robot also traverses through a quarter of the OOI; however, the trajectory went outside of OOI.


%% file: sections/05_conclusion.tex
\section{Conclusion and Future Work}
\label{sec:conclusion}

We presented \textsc{UIVNav}, an underwater navigation method that maximizes collected data of OOI while avoiding obstacles. \textsc{UIVNav} comprises the following steps: (1) creating an intermediate representation (IR) and (2) training the navigation policy using human-labeled IR. The navigation policy remains effective without the need for retraining when there are changes in the domain or OOI. We have demonstrated this by navigating over both oyster and rock reefs. \invis{Our experiments demonstrate the effectiveness of the imitation learning-based approach in providing a solution that, compared to a complete coverage method, will cover 36$\%$ more oysters when traveling the same distance.}Furthermore, our method has been deployed in real-time on BlueROV2 robot in the pool to navigate over the patch of oyster shells. The proposed \textsc{UIVNav} system provides a promising step towards developing generalized and efficient underwater navigation solutions for exploration and monitoring applications. \invis{Moreover, this approach can be adapted to other domains such as forest monitoring or road surveillance applications using aerial vehicles.}

\textbf{Limitations:}
Our goal within this work was to explore limitations of navigation without any localization; therefore, by design \textsc{UIVNav} does not rely on state estimation. As a result, there is no mechanism at the moment to detect the previously visited regions. This can potentially cause robots to visit the same area multiple times without covering the entire region. Nevertheless, our method can be utilized as a local planner for better information gathering within any higher-level global plans when global localization is available.

\textbf{Future Work:}
To demonstrate scalability and applicability in the real world, we plan to conduct extensive open-water surveying operations on real oyster and coral reefs. Additionally, we intend to incorporate the information from the SONAR sensor, specifically range information, for better depth estimation and obstacle detection. The presented work can be also combined with tracking startegies to monitor dynamic interest regions such as school of swimming fish. Finally, we aim to evaluate the versatility of our algorithm by applying it to diverse domains, including both aerial and terrestrial settings.

%% file: paper_template.bbl
\begin{thebibliography}{10}
\providecommand{\url}[1]{#1}
\csname url@rmstyle\endcsname
\providecommand{\newblock}{\relax}
\providecommand{\bibinfo}[2]{#2}
\providecommand\BIBentrySTDinterwordspacing{\spaceskip=0pt\relax}
\providecommand\BIBentryALTinterwordstretchfactor{4}
\providecommand\BIBentryALTinterwordspacing{\spaceskip=\fontdimen2\font plus
\BIBentryALTinterwordstretchfactor\fontdimen3\font minus \fontdimen4\font\relax}
\providecommand\BIBforeignlanguage[2]{{%
\expandafter\ifx\csname l@#1\endcsname\relax
\typeout{** WARNING: IEEEtran.bst: No hyphenation pattern has been}%
\typeout{** loaded for the language `#1'. Using the pattern for}%
\typeout{** the default language instead.}%
\else
\language=\csname l@#1\endcsname
\fi
#2}}

\bibitem{christensen2022recent}
L.~Christensen, J.~de~Gea~Fern{\'a}ndez, M.~Hildebrandt, C.~E.~S. Koch, and B.~Wehbe, ``Recent advances in ai for navigation and control of underwater robots,'' \emph{Current Robotics Reports}, pp. 1--11, 2022.

\bibitem{hansen2018autonomous}
J.~Hansen, S.~Manjanna, A.~Q. Li, I.~Rekleitis, and G.~Dudek, ``Autonomous marine sampling enhanced by strategically deployed drifters in marine flow fields,'' in \emph{OCEANS 2018 MTS/IEEE Charleston}.\hskip 1em plus 0.5em minus 0.4em\relax IEEE, 2018, pp. 1--7.

\bibitem{karapetyan2021meander}
N.~Karapetyan, J.~Moulton, and I.~Rekleitis, ``Meander-based river coverage by an autonomous surface vehicle,'' in \emph{Field and Service Robotics}.\hskip 1em plus 0.5em minus 0.4em\relax Springer, 2021, pp. 353--364.

\bibitem{manjanna2016efficient}
S.~Manjanna, N.~Kakodkar, M.~Meghjani, and G.~Dudek, ``Efficient terrain driven coral coverage using gaussian processes for mosaic synthesis,'' in \emph{2016 13th Conference on Computer and Robot Vision (CRV)}.\hskip 1em plus 0.5em minus 0.4em\relax IEEE, 2016, pp. 448--455.

\bibitem{xanthidis2020navigation}
M.~Xanthidis, N.~Karapetyan, H.~Damron, S.~Rahman, J.~Johnson, A.~O’Connell, J.~M. O’Kane, and I.~Rekleitis, ``Navigation in the presence of obstacles for an agile autonomous underwater vehicle,'' in \emph{2020 IEEE International Conference on Robotics and Automation (ICRA)}.\hskip 1em plus 0.5em minus 0.4em\relax IEEE, 2020, pp. 892--899.

\bibitem{jalal2021underwater}
F.~Jalal and F.~Nasir, ``Underwater navigation, localization and path planning for autonomous vehicles: A review,'' in \emph{2021 International Bhurban Conference on Applied Sciences and Technologies (IBCAST)}.\hskip 1em plus 0.5em minus 0.4em\relax IEEE, 2021, pp. 817--828.

\bibitem{xu2014efficient}
A.~Xu, C.~Viriyasuthee, and I.~Rekleitis, ``Efficient complete coverage of a known arbitrary environment with applications to aerial operations,'' \emph{Autonomous Robots}, vol.~36, pp. 365--381, 2014.

\bibitem{karapetyan2018multi}
N.~Karapetyan, J.~Moulton, J.~S. Lewis, A.~{Quattrini Li}, J.~M. O'Kane, and I.~M. Rekleitis, ``Multi-robot dubins coverage with autonomous surface vehicles,'' in \emph{icra}, 2018.

\bibitem{vidal2019two}
E.~Vidal, N.~Palomeras, K.~Isteni{\v{c}}, J.~D. Hern{\'a}ndez, and M.~Carreras, ``Two-dimensional frontier-based viewpoint generation for exploring and mapping underwater environments,'' \emph{Sensors}, vol.~19, no.~6, p. 1460, 2019.

\bibitem{galceran2015coverage}
E.~Galceran, R.~Campos, N.~Palomeras, D.~Ribas, M.~Carreras, and P.~Ridao, ``Coverage path planning with real-time replanning and surface reconstruction for inspection of three-dimensional underwater structures using autonomous underwater vehicles,'' \emph{Journal of Field Robotics}, vol.~32, no.~7, pp. 952--983, 2015.

\bibitem{xanthidis2021aquavis}
M.~Xanthidis, M.~Kalaitzakis, N.~Karapetyan, J.~Johnson, N.~Vitzilaios, J.~M. O’Kane, and I.~Rekleitis, ``Aquavis: A perception-aware autonomous navigation framework for underwater vehicles,'' in \emph{2021 IEEE/RSJ International Conference on Intelligent Robots and Systems (IROS)}.\hskip 1em plus 0.5em minus 0.4em\relax IEEE, 2021, pp. 5410--5417.

\bibitem{joshi2019experimental}
B.~Joshi, S.~Rahman, M.~Kalaitzakis, B.~Cain, J.~Johnson, M.~Xanthidis, N.~Karapetyan, A.~Hernandez, A.~Q. Li, N.~Vitzilaios, \emph{et~al.}, ``Experimental comparison of open source visual-inertial-based state estimation algorithms in the underwater domain,'' in \emph{2019 IEEE/RSJ International Conference on Intelligent Robots and Systems (IROS)}.\hskip 1em plus 0.5em minus 0.4em\relax IEEE, 2019, pp. 7227--7233.

\bibitem{manderson2018vision}
T.~Manderson, J.~C.~G. Higuera, R.~Cheng, and G.~Dudek, ``Vision-based autonomous underwater swimming in dense coral for combined collision avoidance and target selection,'' in \emph{2018 IEEE/RSJ International Conference on Intelligent Robots and Systems (IROS)}.\hskip 1em plus 0.5em minus 0.4em\relax IEEE, 2018, pp. 1885--1891.

\bibitem{karapetyan2021human}
N.~Karapetyan, J.~V. Johnson, and I.~Rekleitis, ``Human diver-inspired visual navigation: Towards coverage path planning of shipwrecks,'' \emph{Marine Technology Society Journal}, vol.~55, no.~4, pp. 24--32, 2021.

\bibitem{braginsky2016obstacle}
B.~Braginsky and H.~Guterman, ``Obstacle avoidance approaches for autonomous underwater vehicle: Simulation and experimental results,'' \emph{IEEE Journal of oceanic engineering}, vol.~41, no.~4, pp. 882--892, 2016.

\bibitem{yang2021energy}
N.~Yang, D.~Chang, M.~Johnson-Roberson, and J.~Sun, ``Energy-optimal path planning with active flow perception for autonomous underwater vehicles,'' in \emph{2021 IEEE International Conference on Robotics and Automation (ICRA)}.\hskip 1em plus 0.5em minus 0.4em\relax IEEE, 2021, pp. 9928--9934.

\bibitem{perez2016vision}
R.~P{\'e}rez-Alcocer, L.~A. Torres-M{\'e}ndez, E.~Olgu{\'\i}n-D{\'\i}az, and A.~A. Maldonado-Ram{\'\i}rez, ``Vision-based autonomous underwater vehicle navigation in poor visibility conditions using a model-free robust control,'' \emph{Journal of Sensors}, vol. 2016, 2016.

\bibitem{manzanilla2019autonomous}
A.~Manzanilla, S.~Reyes, M.~Garcia, D.~Mercado, and R.~Lozano, ``Autonomous navigation for unmanned underwater vehicles: Real-time experiments using computer vision,'' \emph{IEEE Robotics and Automation Letters}, vol.~4, no.~2, pp. 1351--1356, 2019.

\bibitem{bobkov2016adaptive}
V.~A. Bobkov, V.~Y. Mashentsev, A.~Y. Tolstonogov, and A.~P. Scherbatyuk, ``Adaptive method for auv navigation using stereo vision,'' in \emph{The 26th International Ocean and Polar Engineering Conference}.\hskip 1em plus 0.5em minus 0.4em\relax OnePetro, 2016.

\bibitem{galceran2012efficient}
E.~Galceran and M.~Carreras, ``Efficient seabed coverage path planning for asvs and auvs,'' in \emph{2012 IEEE/RSJ International Conference on Intelligent Robots and Systems}.\hskip 1em plus 0.5em minus 0.4em\relax IEEE, 2012, pp. 88--93.

\bibitem{vidal2017online}
E.~Vidal, J.~D. Hern{\'a}ndez, K.~Isteni{\v{c}}, and M.~Carreras, ``Online view planning for inspecting unexplored underwater structures,'' \emph{IEEE Robotics and Automation Letters}, vol.~2, no.~3, pp. 1436--1443, 2017.

\bibitem{Manderson2020rss}
T.~Manderson, J.~C. Gamboa~Higuera, S.~Wapnick, J.-F. Tremblay, F.~Shkurti, D.~Meger, and G.~Dudek, ``Vision-based goal-conditioned policies for underwater navigation in the presence of obstacles,'' \emph{Robotics: Science and Systems XVI}, Jul 2020.

\bibitem{liu2022learning}
T.~Liu, M.~Yu, and N.~Chopra, ``Learning-based autonomous underwater vehicle navigation following human actions in confined environment,'' in \emph{OCEANS 2022, Hampton Roads}.\hskip 1em plus 0.5em minus 0.4em\relax IEEE, 2022, pp. 1--8.

\bibitem{manderson2020vision}
T.~Manderson, J.~C.~G. Higuera, S.~Wapnick, J.-F. Tremblay, F.~Shkurti, D.~Meger, and G.~Dudek, ``Vision-based goal-conditioned policies for underwater navigation in the presence of obstacles,'' \emph{arXiv preprint arXiv:2006.16235}, 2020.

\bibitem{smolyanskiy2017toward}
N.~Smolyanskiy, A.~Kamenev, J.~Smith, and S.~Birchfield, ``Toward low-flying autonomous mav trail navigation using deep neural networks for environmental awareness,'' in \emph{2017 IEEE/RSJ International Conference on Intelligent Robots and Systems (IROS)}.\hskip 1em plus 0.5em minus 0.4em\relax IEEE, 2017, pp. 4241--4247.

\bibitem{kirillov2023segment}
A.~Kirillov, E.~Mintun, N.~Ravi, H.~Mao, C.~Rolland, L.~Gustafson, T.~Xiao, S.~Whitehead, A.~C. Berg, W.-Y. Lo, \emph{et~al.}, ``Segment anything,'' \emph{arXiv preprint arXiv:2304.02643}, 2023.

\bibitem{lin2023oysternet}
X.~Lin, N.~J. Sanket, N.~Karapetyan, and Y.~Aloimonos, ``Oysternet: Enhanced oyster detection using simulation,'' in \emph{2023 IEEE International Conference on Robotics and Automation (ICRA)}.\hskip 1em plus 0.5em minus 0.4em\relax IEEE, 2023, pp. 5170--5176.

\bibitem{ronneberger2015u}
O.~Ronneberger, P.~Fischer, and T.~Brox, ``U-net: Convolutional networks for biomedical image segmentation,'' in \emph{Medical Image Computing and Computer-Assisted Intervention--MICCAI 2015: 18th International Conference, Munich, Germany, October 5-9, 2015, Proceedings, Part III 18}.\hskip 1em plus 0.5em minus 0.4em\relax Springer, 2015, pp. 234--241.

\bibitem{yu2023udepth}
B.~Yu, J.~Wu, and M.~J. Islam, ``Udepth: Fast monocular depth estimation for visually-guided underwater robots,'' in \emph{2023 IEEE International Conference on Robotics and Automation (ICRA)}.\hskip 1em plus 0.5em minus 0.4em\relax IEEE, 2023, pp. 3116--3123.

\bibitem{lin2022oystersim}
X.~Lin, N.~Jha, M.~Joshi, N.~Karapetyan, Y.~Aloimonos, and M.~Yu, ``Oystersim: Underwater simulation for enhancing oyster reef monitoring,'' in \emph{OCEANS 2022, Hampton Roads}.\hskip 1em plus 0.5em minus 0.4em\relax IEEE, 2022, pp. 1--6.

\bibitem{chollet2015keras}
\BIBentryALTinterwordspacing
F.~Chollet \emph{et~al.} (2015) Keras. [Online]. Available: \url{https://github.com/fchollet/keras}
\BIBentrySTDinterwordspacing

\bibitem{laina2016deeper}
I.~Laina, C.~Rupprecht, V.~Belagiannis, F.~Tombari, and N.~Navab, ``Deeper depth prediction with fully convolutional residual networks,'' in \emph{2016 Fourth international conference on 3D vision (3DV)}.\hskip 1em plus 0.5em minus 0.4em\relax IEEE, 2016, pp. 239--248.

\bibitem{Hunter:2007}
J.~D. Hunter, ``Matplotlib: A 2d graphics environment,'' \emph{Computing in Science \& Engineering}, vol.~9, no.~3, pp. 90--95, 2007.

\bibitem{joyce2011kullback}
J.~M. Joyce, ``Kullback-leibler divergence,'' in \emph{International encyclopedia of statistical science}.\hskip 1em plus 0.5em minus 0.4em\relax Springer, 2011, pp. 720--722.

\bibitem{chow2009brownian}
W.~C. Chow, ``Brownian bridge,'' \emph{Wiley interdisciplinary reviews: computational statistics}, vol.~1, no.~3, pp. 325--332, 2009.

\bibitem{horne2007analyzing}
J.~S. Horne, E.~O. Garton, S.~M. Krone, and J.~S. Lewis, ``Analyzing animal movements using brownian bridges,'' \emph{Ecology}, vol.~88, no.~9, pp. 2354--2363, 2007.

\bibitem{pang2019swarm}
B.~Pang, Y.~Song, C.~Zhang, H.~Wang, and R.~Yang, ``A swarm robotic exploration strategy based on an improved random walk method,'' \emph{Journal of Robotics}, vol. 2019, pp. 1--9, 2019.

\bibitem{wagner1998robotic}
I.~A. Wagner, M.~Lindenbaum, and A.~M. Bruckstein, ``Robotic exploration, brownian motion and electrical resistance,'' in \emph{Randomization and Approximation Techniques in Computer Science: Second International Workshop, RANDOM’98 Barcelona, Spain, October 8--10, 1998 Proceedings 2}.\hskip 1em plus 0.5em minus 0.4em\relax Springer, 1998, pp. 116--130.

\bibitem{blender}
\BIBentryALTinterwordspacing
B.~O. Community, \emph{Blender - a 3D modelling and rendering package}, Blender Foundation, Stichting Blender Foundation, Amsterdam, 2018. [Online]. Available: \url{http://www.blender.org}
\BIBentrySTDinterwordspacing

\bibitem{robotics2016bluerov2}
B.~Robotics, ``Bluerov2: The world’s most affordable high-performance rov,'' \emph{BlueROV2 Datasheet; Blue Robotics: Torrance, CA, USA}, 2016.

\end{thebibliography}
